\documentclass[final]{article}
\pdfoutput=1
\PassOptionsToPackage{numbers}{natbib}
\usepackage[]{nips_2018}

\usepackage[utf8]{inputenc} % allow utf-8 input
\usepackage[T1]{fontenc}    % use 8-bit T1 fonts
\usepackage{booktabs}       % professional-quality tables
\usepackage{amsfonts}       % blackboard math symbols
\usepackage{nicefrac}       % compact symbols for 1/2, etc.
\usepackage{microtype}      % microtypography
\usepackage{mathtools}
\DeclarePairedDelimiter{\ceil}{\lceil}{\rceil}

\usepackage{graphicx}
\usepackage{amsmath,amssymb} % define this before the line numbering.
\usepackage{color}
\usepackage{xcolor}
\definecolor{hcitecolor}{RGB}{40,180,40}
\usepackage[colorlinks=true,citecolor=hcitecolor, linkcolor=blue]{hyperref}
\usepackage{paralist}
\usepackage{tikz}
\usetikzlibrary{shadows,matrix, calc, positioning, arrows,shapes,backgrounds, arrows.meta, quotes,patterns,snakes}
\usepackage{rbase}

\newcommand{\framedef}[1]{\centerline{\fbox{~~\parbox{.96\textwidth}{\vspace{2pt} #1 \vspace{5pt}}~~}}}

\newtheorem{theorem}{Theorem}

\newtheorem{prop}[theorem]{Proposition}
\newtheorem{lemma}[theorem]{Lemma}

\newtheorem{defn}{Definition}

\newcommand{\sm}[1]{\m{\smash{#1}}}
\newcommand{\nnl}{s}
\newcommand{\fell}{f^{\nnl}}
\newcommand{\fellm}{f^{\nnl-1}}
\newcommand{\chiell}{h^{\nnl}}

\title{Clebsch--Gordan Nets: a Fully Fourier Space Spherical Convolutional Neural Network}

\author{
	Risi Kondor$^1$\thanks{Authors are arranged alphabetically} \quad Zhen Lin$^1$\footnotemark[1] \quad Shubhendu Trivedi$^2$\footnotemark[1] \\
	$^1$The University of Chicago \quad  $^2$Toyota Technological Institute\\
	\texttt{\{risi, zlin7\}@uchicago.edu, shubhendu@ttic.edu}\\	
}

\begin{document}
% \nipsfinalcopy is no longer used

\maketitle

\begin{abstract}
Recent work by Cohen \emph{et al.} \cite{CohenSpherical18} has achieved state-of-the-art results for learning spherical 
images in a rotation invariant way by using ideas from group representation theory and 
noncommutative harmonic analysis. In this paper we propose a generalization of this work 
that generally exhibits improved performace, but from an implementation point of view is 
actually simpler. 
An unusual feature of the proposed architecture is that it uses the Clebsch--Gordan transform 
as its only source of nonlinearity, thus avoiding repeated forward and backward Fourier transforms. 
The underlying ideas of the paper generalize to constructing neural networks that are invariant 
to the action of other compact groups. 
\end{abstract}

\section{Introduction}

Despite the many recent breakthroughs in deep learning, we still do not have a satisfactory understanding of 
how deep neural networks are able to achieve such spectacular perfomance 
on a wide range of learning problems. 
One thing that is clear, however, is that certain architectures pick up on 
natural invariances in data, and this is a key component to their success. 
%Nowhere is this more clear than in the case of classical 
The classic example is of course Convolutional Neural Networks (CNNs) 
for image classification \cite{LeCun1989}.  
Recall that, fundamentally, each layer of a CNN realizes two simple operations: 
a linear one consisting of convolving the previous layer's activations with a (typically small) 
learnable filter, and a nonlinear but pointwise one, such as a ReLU operator\footnote{Real 
CNNs typically of course have multiple channels, and correspondingly multiple filters per layer,  
%the channels might also be allowed to mix with each other, 
but this does not fundamentally change the network's invariance properties.}. 
This architecture is sufficient to guarantee \emph{translation equivariance}, 
meaning that if the input image is translated by some vector \m{\V t}, 
then the activation pattern in each higher layer of the network will 
%remain the same, except for being 
translate by the same amount. 
Equivariance is crucial to image recognition for two closely related reasons: 
(a) It guarantees that exactly the same filters are  
applied to each part the input image regardless of position. 
(b) Assuming that finally, at the very 
top of the network, we add some layer that is translation \emph{invariant}, the entire network will 
be invariant, ensuring that it can detect any given object equally well regardless of its location.   

Recently, a number of papers have appeared that examine equivariance from the theoretical point of view, 
motivated by the understanding that the natural way to generalize convolutional networks to 
other types of data will likely lead through generalizing the notion of equivariance itself to 
other transformation groups \cite{Gens2014,Cohen2016,Cohen2017,Poczos2017,EquivarianceArxiv18}. 
Letting \sm{f^s} denote the activations of the neurons in layer \m{s} of a hypothetical generalized 
convolution-like neural network, 
mathematically, equivariance to a group \m{G} means that if the inputs to the network are transformed 
by some transformation \m{g\tin G}, then \sm{f^s} transforms to \m{T^s_g(f^s)} for 
some fixed set of linear transformations \sm{\cbrN{T^s_g}_{g\in G}}. 
%It should be noted that in this generalized setting some authors use the word 
(Note that in some contexts this is called ``covariance'', the difference between 
the two words being only one of emphasis.)
% and we will sometimes use the two words interchangeably.) 

\ignore{
The linearity assumption on the \m{T^\ell_g} transformations is a fairly mild and natural 
condition. From the mathematical point of view, however, it has crucial consequences, because 
it means that  \sm{\cbrN{T^\ell_g}_{g\in G}} form what is known as a \emph{representation} of the group \m{G}. 
Representation theory is a well developed branch of algebra offering a wealth of powerful 
results to lean on. 
In particular, representation theoretic ideas can help inform the design of 
neural networks by prescribing what form each layer must take to attain equivariance.   
}

A recent major success of this approach are Spherical CNNs \cite{CohenSpherical18}\cite{EstevesSph}, 
which are an \m{\SO(3)}--equivariant neural network 
architecture for learning images painted on the sphere\footnote{\m{\SO(3)} denotes the group of three 
dimensional rotations, i.e., the group of \m{3\<\times 3} orthogonal matrices.}.  %in a rotation invariant way, 
Learning images on the sphere in a rotation invariant way has  applications in a wide range of domains 
from 360 degree video through drone navigation to molecular 
chemistry \cite{Zelnik-Manor2005,Cruz-Mota2012,Su2017a,Lai,Khasanova2017,Boomsma2017,Su2017}. 
The key idea in Spherical CNNs is to generalize convolutions %to the sphere 
using the machinery of noncommutative harmonic analysis: 
employing a type of generalized \m{\SO(3)} Fourier transform \cite{Healy96,Kostelec2008},  
Spherical CNNs transform the image 
to a sequence of matrices, and compute the spherical analog of convolution in Fourier space. 
This beautiful construction guarantees equivariance, 
and the resulting network attains state of the art results on several benchmark datasets. 

One potential drawback of Spherical CNNs of the form proposed in \cite{CohenSpherical18}, however,    
is that the nonlinear transform in each layer still needs to be computed in ``real space''. 
Consequently, each layer of the network involves a forward and a backward \m{\SO(3)} Fourier transform, 
%transform back and forth between real space and Fourier space, 
which is relatively costly, and is a source of numerical errors, especially since 
the sphere and the rotation group do not admit any regular discretization similar to the square grid 
for Euclidean space. 

%Spherical CNNs start by transforming the image on the sphere with a type of  
%generalized \m{\SO(3)} Fourier transform, and actually compute the convolution in Fourier space. 
%This is a fundamentally representation theoretic approach, because the generalized \m{\SO(3)} 
%Fourier transform is defined in terms of the irreducible representations of \m{\SO(3)}. 

Spherical CNNs are not the only context in which the idea of Fourier space neural networks 
has recently appeared \cite{Worrall2016,Esteves,Cohen2017,EquivarianceArxiv18}. 
From a mathematical point of view, the relevance of Fourier theoretic ideas in 
all these cases is a direct consequence of equivariance, specifically, of the fact that the 
\sm{\cbrN{T^s_g}_{g\in G}} operators form a \emph{representation} of the underlying group, 
in the algebraic sense of the word \cite{Serre}. 
%It can be shown 
%Such ideas can also be applied much more broadly, to create neural networks that are equivariant to 
%the action of \emph{any} compact group, not just rotations. 
%In another recent paper, it is shown 
In particular, it has been shown 
that whenever there is a compact group \m{G} acting on the inputs 
of a neural network, there is a natural notion of Fourier transformation with respect to \m{G}, yielding 
a sequence of Fourier matrices \sm{\cbrN{F^s_\ell}_{\ell}} at each layer, and the 
linear operation at layer \m{s} will be equivariant to \m{G} if and only if it is equivalent to 
multiplying each of these matrices from the right by some (learnable) filter matrix \sm{H^s_\ell} 
\cite{EquivarianceArxiv18}. 
Any other sort of operation will break equivariance. 
The spherical convolutions employed in \cite{CohenSpherical18} 
are a special case of this general setup for \m{\SO(3)}, and 
the ordinary convolutions employed in classical CNNs are a special case for the integer 
translation group \m{\ZZ^2}.  
In all of these cases, however, the issue remains that the nonlinearities need to be computed 
in ``real space'', necessitating repeated forward and backward Fourier transforms.  

In the present paper we propose a spherical CNN that differs from \cite{CohenSpherical18} 
in two fundamental ways: 
\begin{compactenum}[~~1.]
\item While retaining the connection to noncommutative Fourier analysis, we relax the requirement 
that the activation of each layer of the network needs to be a 
(vector valued) function on \m{\SO(3)}, requiring only that it be expressible as a collection 
of some number of \m{\SO(3)}--covariant vectors (which we call fragments) corresponding to different irreducible 
representations of the group. In this sense, our architecture is strictly more general than 
\cite{CohenSpherical18}. 
\item Rather than a pointwise nonlinearity in real space, our network takes the tensor (Kronecker) product 
of the activations in each layer followed by decomposing the result into irreducible fragments 
using the so-called \emph{Clebsch--Gordan decomposition}. 
This way, we get a ``fully Fourier space'' neural network that avoids repeated forward and 
backward Fourier transforms.  
%needing just a single Fourier transform of the input data in layer zero.  
\end{compactenum}
The resulting architecture is not only more flexible and easier to implement than \cite{CohenSpherical18}, 
but our experiments show that it can also perform better on some standard datasets. 

The Clebsch--Gordan transform has recently appeared in two separate preprints discussing  
neural networks for learning physical systems \citep{ThomasTensor18,KondorNbodyArxiv18}. 
However, to the best of our knowledge, it has never been proposed as a general purpose 
nonlinearity for covariant neural networks. 
In fact, any compact group has a Clebsch--Gordan decomposition (although, due to its connection to 
angular momentum in physics, the \m{\SO(3)} case is by far the best known), so, in principle, the 
methods of the present paper could be applied much broadly, in any situation where one desires to 
build a neural network that is equivariant to some class of transformations captured by a compact group. 

%can be defined for any compact group, but of course its form will 
%In this sense, the message of the present paper goes significantly beyond just learning on the sphere. 
%depend on the group, and it so happens that because of its relevance to angular momentum in physics, 
%the \m{\SO(3)} Clebsch--Gordan transform is the best known case. 

\ignore{
All this make a compelling case for Fourier space architectures in generalized 
equivariant neural networks. 
However, in addition to the linear part of each layer's operation, the nonlinearity also has to be 
equivariant to the action of \m{G}, and all existing equivariant neural nets achieve that by making 
it a pointwise nonlinearity, such as a ReLU, in ``real space''. 
Consequently, Fourier space networks have to shuttle back and forth between real space and Fourier space, 
which is computationally expensive and potentially also involves a degradation in accuracy (and equivariance). 
In the case of Spherical CNNs, for example, 
%the forward transform essentially reduces to a collection of 
%spherical harmonic transforms, while 
the backward transform is essentially a collection of inverse spherical harmonic transform 
that projects the activations back onto the sphere, or, more accuractely, \m{\SO(3)}. 
However, the sphere and \m{\SO(3)} do not have completely regular discretizations like the Eucliean 
plane does. Therefore, the inverse Fourier transform inevitably involves quadrature and 
some discretization error.  

In the present paper we propose a neural network architecture where the nonlinearity is a 
\emph{Clebsch--Gordan transform}, which has the advantage of being an operation that is 
both covariant to the group action, and computable in Fourier space. 
Thus, to the best of our knowledge, our architecture is the first ``fully Fourier space'' covariant neural 
network. 
In the following pages we develop Clebsch--Gordan networks for the specific case of rotation equivariant 
image recognition on the sphere, and show that on the some standard datasets it can yield 
better results than . 
}

%\textbf{Related work} 

\section{Convolutions on the sphere}

%Convolutional neural networks (CNNs) have proved to be extraordinarily successful at a range of Computer Vision tasks 
%from object recognition to video segmentation \cite{LeCun1989,Krizhevsky2012,He2016}. 
The simplest example of a covariant neural network is a classical \m{S\<+1} layer CNN for image recognition.  
In each layer of a CNN the neurons are arranged in a rectangular grid, so (assuming for simplicity 
that the network has just one channel) the activation of layer \m{s} can be regarded as a function 
\m{\fell\colon \ZZ^2\to\RR}, with \m{f^0} being the input image.  
%with just a single channel, the output \m{\fell(\x)} of layer number \m{\nnl} is %in a classical CNN is 
The neurons compute \m{\fell} by taking the cross-correlation\footnote{Convolution and cross-correlation are closely related mathematical 
concepts that are somewhat confounded in the deep learning literature. 
In this paper we are going to be a little more precise and focus on cross-correlation, because, 
despite their name, that is what CNNs actually compute.}
of the previous layer's output \ignore{\fellm}with a small (learnable) filter \m{\chiell},
\begin{equation}\label{eq: xcorr1}
(\chiell\<\star \fellm)(x)=
\sum_{y}\;\,\chiell(y\<-x)\;\fellm(y),
\end{equation}
and then applying a nonlinearity \m{\sigma}, such as the Re-LU operator:
\begin{equation}\label{eq: CNN1}
\fell(x)=\sigma\nts\brN{(\chiell\<\star \fellm)(x)}.
\end{equation}
%Generally, \m{\chiell} has small support (often just \m{3\<\times 3} or \m{5\<\times 5} pixels in size), and it is these 
%filters that the network learns from training data. 
%As Cohen \emph{et al.} point out in \cite{SphericalCNN2018}, 
%An alternative interpretation emerges when we realize that 
Defining\: \sm{T_x(\chiell)(y)=\chiell(y-x)}, which is  
nothing but \m{\chiell} translated by \m{x}, allows us to equivalently write  
%This allows us to rewrite 
\rf{eq: xcorr1} as %in the form of an inner product 
%However, the right hand side of \rf{eq: xcorr1} can eqivalently be interpreted as an inner product 
\begin{equation}\label{eq: xcorr2}
(\chiell\<\star \fellm)(x)=\inpN{\fellm,T_x(\chiell)}, 
%=\sum_{\y} f^{s-1}(\y)\:\chi_s^{(\x)}(\y). 
\end{equation}
where the inner product is \m{\inpN{\fellm\!,T_x(\chiell)}\<=\sum_{y} \fellm(y)\,T_x(\chiell)(y). }
%\sm{\inp{g,h}=\sum_{\y}{g(\y) h(\y')}} 
%between \m{\fellm} and \sm{[\chi^s]^{\x}(\y)=\chi}
%and the filter \sm{\chiell} \emph{after} \sm{\chiell} has been translated by \m{\x}. 
%Thus, 
What this formula tells us is that fundamentally each layer of the CNN just does pattern matching: 
\m{\fell(x)} is an indication of how well the part of \sm{\fellm} around \sm{x} matches the filter \sm{\chiell}. 
% matches the 

Equation \ref{eq: xcorr2} is the natural starting point for generalizing convolution to 
the unit sphere, \m{S^2}.  
An immediate complication that we face, however, is that unlike the plane, \m{S^2} cannot be discretized 
by any regular (by which we mean rotation invariant) arrangement of points. 
A number of authors have addressed this problem in different ways \cite{Boomsma2017,Su2017}. 
Instead of following one of these approaches, similarly to recent work on manifold CNNs 
\cite{Masci2015,Monti2016}, 
in the following we simply treat each \m{\fell} and the corresponding filter \m{\chiell} 
as continuous functions on the sphere, \m{\fell(\theta,\phi)} and \m{\chiell(\theta,\phi)}, 
where \m{\theta} and \m{\phi} are the polar and azimuthal angles. 
We allow both these functions to be complex valued, the reason for which will become clear later. 

The inner product of two complex valued functions on the surface of the sphere is given by the formula 
\begin{equation}\label{eq: inp}
\inp{g,h}_{S_2}=
%\int_{S_2} \!\!g^\ast(\x)\tts h(\x)\,d\mu(\x)=
\ovr{4\pi}\int_{0}^{2\pi} \int_{-\pi}^{\pi} \sqbN{g(\theta,\phi)}^\ast \ts h(\theta,\phi)\tts
\cos\theta\ts d\theta\ts d\phi, 
\end{equation} 
where \m{{}^\ast} denotes complex conjugation. 
Further, \m{h} (dropping the layer index for clarity) 
can be moved to any point \m{(\theta_0,\phi_0)} on \m{S^2} by taking 
\sm{h'(\theta,\phi)=h(\theta\<-\theta_0,\phi\<-\phi_0)}. 
This suggests that the generalization of \ref{eq: xcorr2} to the sphere should be 
\begin{equation}\label{eq: S^2 convo0}
(h\star f)(\theta_0,\phi_0)=
\ovr{4\pi}\int_{0}^{2\pi} \int_{-\pi}^{\pi} \sqbN{h(\theta\<-\theta_0,\phi\<-\phi_0)}^\ast\ts 
f(\theta,\phi)\tts\cos\theta\ts d\theta\ts d\phi. 
%\int_{S_2}\chi\brbig{(R^{(z)}_{\phi })^{-1}(R^{(y)}_\theta}^{-1} x)^\ast\:f(x)\,d\mu(\x). 
%\ovr{4\pi}\int_{0}^{2\pi} \int_{-\pi}^{\pi} f^\ast(\theta,\phi)\,g(\theta,\phi)\,\cos\theta\,d\theta d\phi.
\end{equation}
Unfortunately, this generalization would be \emph{wrong}, because it does not take into account that 
\m{h} can also be rotated around a third axis.
%, namely the ``x'' axis in Figure \ref{fig: convo1}.\input{fig-convo1}
\ignore{
Another way to see this problem is by considering the different ways that \m{h} can be 
moved to the same position on the sphere. For example, \m{h} can be moved to the North pole 
by a single rotation \sm{T^{(y)}_{\pi/2}} of angle \m{\pi/2} about the \m{y} axis. 
It can also be moved to the same location by a rotation \sm{T^{(z)}_{\pi/2}} about the \m{z} axis followed 
by a rotation \sm{T^{(x)}_{\pi/2}} about the \m{x} axis. While these two options move \m{h} to the same 
location, the \emph{orientation} of the two rotated images will differ by an angle of \m{\pi/2}. 
This argument demonstrates that one fundamental difference between classical CNNs and the spherical case is 
that rotation group underlying the latter is not commutative. 
}
The correct way to generalize cross-correlations to the sphere is to define \m{h\<\star f} as a function 
\emph{on the rotation group itself}, i.e., to set 
\begin{equation}\label{eq: S^2 xcorr}
(h\star f)(R)=
%\int_{\SO(3)}T_R(h)(\theta,\^\ast f(\x)\,d\mu(\x), 
\ovr{4\pi}\int_{0}^{2\pi} \int_{-\pi}^{\pi} \sqbbig{h_R(\theta,\phi)}^\ast\,
f(\theta,\phi)\,\cos\theta\,d\theta\, d\phi
\qqquad R\tin\SO(3),
\end{equation}
where \m{h_R} is \m{h} rotated by \m{R}, expressible as 
%Depending on whether we choose to parametrize \m{\SO(3)} in terms of Euler angles 
%\m{(\alpha,\beta,\gamma)} 
%or rotation matrices, \m{h_R} can be expressed as 
%the translate of \m{h} by \m{R} is defined 
\begin{equation}\label{eq: rotation}
%h_R(\theta,\phi)=\brN{T^{(y)}_\gamma T^{(x)}_\beta T^{(z)}_\alpha}(h)(\theta,\phi) 
%\qqquad\text{or} \qqquad 
h_R(x)=h(R^{-1}x),
\end{equation}
with \m{x} being the point on the sphere at position \m{(\theta,\phi)} (c.f. \cite{Gutman2008}\cite{CohenSpherical18}). 

\subsection{Fourier space filters and activations}

Cohen et al.\cite{CohenSpherical18} observe that the double integral in \rf{eq: S^2 xcorr} 
would be extremely inconvenient to compute in a neural network. 
As mentioned, in the case of the sphere, 
just finding the right discretizations to represent \m{f} and \m{h} is already problematic.  
As an alternative, it is natural to represent both these functions in terms of their 
spherical harmonic expansions 
\begin{equation}\label{eq: sh expansion}
f(\theta,\phi)= \sum_{\ell=0}^{\infty} \sum_{m=-\ell}^{\ell} \h{f}_\ell^m\:Y_\ell^m(\theta,\phi)
\qqquad   
h(\theta,\phi)= \sum_{\ell=0}^{\infty} \sum_{m=-\ell}^{\ell} \h{h}_\ell^m\:Y_\ell^m(\theta,\phi). 
\end{equation}
Here, \m{Y_\ell^m(\theta,\phi)} are the well known spherical harmonic functions indexed by 
\m{\ell=0,1,2,\ldots} and \m{m\tin\cbrN{-\ell,-\ell\<+1,\ldots,\ell}}. 
The spherical harmonics form an orthonormal basis for \m{L_2(S^2)}, so 
\rf{eq: sh expansion} can be seen as a kind of Fourier series on the sphere,  
%setting a resolution limit \m{L}, 
in particular, the elements of the \m{f_0,f_1,f_2,\ldots} 
%\m{\brN{f_\ell\<=(f_\ell^{-\ell},\ldots,f_\ell^\ell)}_{\ell=0}^L} 
%and \m{\brN{h_\ell=(h_\ell^{-\ell},\ldots,h_\ell^\ell)}_{\ell=0}^L} 
coefficient vectors can be computed relatively easily by 
%are given by %can easily be computed via 
\begin{equation*}
\h{f}_\ell^m=\ovr{4\pi}\int_{0}^{2\pi} \int_{-\pi}^{\pi} f(\theta,\phi)\,Y_\ell^m(\theta,\phi) 
\,\cos\theta\,d\theta\, d\phi, 
\end{equation*}
and similarly for \m{h}. 
Similarly to usual Fourier series, in practical scenarios spherical harmonic expansions 
are computed up to some limiting ``frequency'' \m{L}, which depends on the desired resolution.  
%in represent our signal. 

Noncommutative harmonic analysis \cite{Terras,Diaconis} tells us that 
functions on the rotation group also admit a type of generalized Fourier transform.  
Given a function \m{g\colon \SO(3)\to\CC}, the Fourier transform of \m{g} is defined as the collection 
of \emph{matrices} 
\begin{equation}
G_\ell=\ovr{4\pi}\int_{\SO(3)} g(R)\,\rho_\ell(R)\ts d\mu(R) 
%\,\cos\theta\,d\theta\, d\phi 
\hspace{50pt}\ell=0,1,2,\ldots,
\end{equation}
where \m{\rho_\ell\colon \SO(3)\to\CC^{(2\ell+1)\times(2\ell+1)}} 
are fixed matrix valued functions called the irreducible representations of \m{\SO(3)}, 
sometimes also called Wigner D-matrices.  
Here \m{\mu} is a fixed measure called the Haar measure that just hides factors similar to the 
\m{\cos\theta} appearing in \rf{eq: inp}. 
For future reference we also note that one dimensional irreducible representation \m{\rho_0} 
is the constant representation \sm{\rho_0(R)=(1)}. 
The inverse Fourier transform is given by 
\begin{equation*}
g(R)=\sum_{\ell=0}^{\infty}\tr\sqb{G_\ell\, \rho_\ell(R^{-1})}\hspace{80pt}R\tin\SO(3).
\end{equation*}
While the spherical harmonics can be chosen to be real, the \m{\rho_\ell(R)} representation matrices 
are inherently complex valued. This is the reason that we allow all other quantities, including 
the \m{\fell} activations and \m{\chiell} filters to be complex, too. 

Remarkably, the above notions of harmonic analysis on the sphere and the rotation group 
are closely related. 
In particular, it is possible to show that each Fourier component of the  
spherical cross correlation \rf{eq: S^2 xcorr} that we are interested in computing is given simply by 
the outer product %of the corresponding \m{\wt{f}_\ell} and \m{\wt{h}_\ell^\top} vectors:
\begin{equation}\label{eq: xcorr Fourier}
\sqbN{\h{h\<\star f}}_\ell=\h{f}_\ell \cdot \h{h}_\ell^\dag
\hspace{80pt} \ell=0,1,2,\ldots,L,
\end{equation}
where \m{{}^\dag} denotes the conjugate transpose (Hermitian conjugate) operation. 
Cohen et al.'s Spherical CNNs \cite{CohenSpherical18}  are essentially based on this formula. 
In particular, they argue that instead of the continuous function \m{f}, it is more expedient 
to regard the components of the \sm{\seqz{\h{f}}{L}} vectors as the ``activations'' of 
their neural network, while the learnable weights or filters are the 
\sm{\seqz{\h{h}}{L}} vectors. 
Computing spherical convolutions in Fourier space then reduces to just computing 
a few outer products. 
Layers \m{s=2,3,\ldots,S} of the Spherical CNN operate similarly, except that %the input activation 
\m{\fellm} is a function on \m{\SO(3)}, so \rf{eq: S^2 xcorr} must be replaced by 
cross-correlation on \m{\SO(3)} itself, and \m{h} must also be a function on \m{\SO(3)} rather than 
just the sphere. 
Fortuitiously, the resulting cross-correlation formula is almost exactly the same: 
\begin{equation}\label{eq: xcorr Fourier2}
\sqbN{\h{h\<\star f}}_\ell=F_\ell \cdot H_\ell^\dag
\hspace{100pt} \ell=0,1,2,\ldots,L,
\end{equation}
apart from the fact that now \m{F_\ell} and \m{H_\ell} are matrices (see \cite{CohenSpherical18} for details). 
%which fortuitously turns out to be of the same form 
%as \rf{eq: xcorr Fourier}, except that \sm{\wt{f}_\ell} and \sm{\wt{h}_\ell} 
%are now replaced by the corresponding \m{F_\ell} and \m{H_\ell}  
%}matrices (see \cite{CohenSpherical18} for details). 

%which form an or
%\begin{equation}
%Y_\ell^m(\theta,\phi)=\sqrt{\fr{2\ell\<+1}{4\pi}\fr{(\ell\<-m)!}{(\ell+m)!}}\: 
%P_\ell^m(\cos \theta)\,e^{\iota m\phi}
%\qquad ... -\ell\leq m\leq \ell\qquad \ell=0,1,\ldots
%\end{equation} l
%are the well known spherical harmonic functions, and \m{P_\ell^m} are the associated Legendre polynomials. 

% ------------------------------------------------------------------------

\ignore{
In \cite{SphericalCNN2018} it is pointed out that the correct solution to both the above problems is 
to define the cross-correlation of signals on the sphere as a function %\m{f\ast g\colon S^2\to\CC}, 
on the rotation group \m{\SO(3)} rather than a function on the sphere. Specifically, 
where now \m{R\tin \SO(3)}, i.e., \m{R} is a three dimensional orthogonal matrix. 
% and \m{\mu} is the invariant measure of the group. 
The following section explains how this forbidding looking integral can be computed 
efficiently by working in Fourier space. 
}

\ignore{
given by a formula such as 
%the activation (output) of neuron 
%\m{\neuron^\ell_{\x}} is given by %in a CNN is computed by first computing the cross-correa
%a convolution followed by a nonlinearity \m{\sigma}: 
\begin{equation}\label{eq: convo1}
f^\ell(x_1,x_2)=\sigma\brBig{b+
\sum_{y_1=-w_1}^{w_1} \sum_{y_2=-w_2}^{w_2} f^{\ell-1}(x_1\<+y_1,\ts x_2\<+y_2)\,\chi^\ell(y_1,y_2)}.
\end{equation}
where \m{\chi^\ell} is a learned \m{(2w_1+1)\times (2w_2+1)} filter of weights 
%that is shared across all neurons in the layer 
and \m{\sigma} is a nonlinearity. 
%It is these filters that the network learns from training data. 
}
\ignore{
%Technically, in the mathematics literature \rf{eq: convo1} is called the cross-correlation of \m{f^\ell} 
%with \m{\chi_\ell} and denoted \m{f^{\ell-1}\<\odot \chi^\ell}. Convolution is a closely related concept, so, following 
%the neural networks convention, we refer to both as convolution type operations. 
What is crucial about cross-correlation 
is that they reflect the underlying symmetries 
of the problem domain. In particular, extending \m{\chi} to the entirety of the unit grid \m{\ZZ^2} 
on the plane by padding it with zeros outside of the \m{(2w_1\<+1)\times (2w_2\<+1)} rectangle, 
\rf{eq: convo1} can be written as 
where \m{T_{\x}(\y)=\x\<+\y}, which is the so-called \emph{action} of translation by \m{\x} on the plane. 
Rewriting the convolution formula in this way makes it clear is that convolution is a type of template 
matching: \m{f^{\ell-1}\odot \chi^\ell} 
expresses what the value of the inner product between \m{f^{\ell-1}} and 
\m{\chi^\ell} is \emph{after} \m{f^{\ell-1}} has been translated by \m{\x}. 
As pointed out in \cite{SphericalCNN2018}, this formula is the key to generalizing 
convolution to other domains, such as the sphere. 

%As emphasized by \cite{SphericalCNN2018}, \rf{eq: convo2} is crucial for generalizing the convolutional 
%neural networks to more general domains, including the sphere. 

It is widely recognized that the main reason for the success of CNNs is their intimate 
relationship with the nature of translations. In particular, CNNs satisfy a property called \m{equivariance}, 
which we define formally as follows. 

\begin{definition}
Let \m{\Ncal} be an \m{L\<+1} layer neural network. 
Assume that there is a group \m{G} that acts on each layer by an action \m{T^\ell}. We say that \m{\Ncal} is \emph{equivariant} with respect to \m{G} if when the inputs to the neural 
network are transformed \m{f^0\mapsto {f^0}'} where \m{{f^0}'(x)=f^0(T_g^{-1}(x))}, then the 
activation of every other layer correspondingly transforms as  
\begin{equation}
f^\ell\mapsto {f^\ell}'\hspace{80pt} {f^\ell}'(x)=f^\ell(T_g^{-1}(x)).
\end{equation}  
\end{definition}   
\smallskip

\noindent
}
 
%Extending both \m{f^{\ell-1}} and the filter \m{\chi^\ell} to the entire plane, Cohen and Welling \cite{SphericalCNN2018} 
%note that \rf{eq: xcorr1} can be written as 
%\begin{equation}\label{eq: convo2}
%(\chi\star f^{\ell-1})(\x)=\sum_{\y} \chi^{\ell}(T^{-1}_\x(\y))\:f^{\ell-1}(\y),
%(\chi\star f^{\ell-1})(\x)=\sum_{\y} \chi^{\ell}(\y-\x)\:f^{\ell-1}(\y), 
%\end{equation}
%where \m{T_{\x}(\y)=\x\<+\y} expresses the effect of translating the plane by \m{\x} units. 
%and hence \m{\chi^{\ell}(T^{-1}_\x(\y))=\chi^\ell(\y-\x)} is \m{\chi^\ell} translated by \m{\x}.   
%The significance of 
%These formulae 
%making it clear that cross-correlation is intimately tied 
%to the structure of the underlying space. 
%Computing \rf{eq: S^2 convo1} explicitly would be cumbersome and expensive for several 
%reasons, 
%including the fact that it would require us to discretize \m{\SO(3)} and also deal with the 
%measure \m{\mu}. 
%Fortunately, as \cite{SphericalCNN2018} explain, using the machinery of noncommutative Fourier analysis, 
%this operation can be greatly simplified. 
%This is what we explain in the following section. 

\ignore{
The well known fact that CNNs are equivariant to translation is closely related. 
\renewcommand{\bt}{\mathbf{t}} Recall that equivariance means that if we translate the inputs of a %convolutional neural network 
CNN by some amount \m{\bt}, 
\begin{equation}
f^0\mapsto f'{}^0 \hspace{90pt} f'{}^0(\x)=f^0(\x-\bt),
\end{equation}
then the activations of the higher layers \m{\nnl=1,2,\ldots,L} transform the same way, %i.e., 
\begin{equation}
\fell\mapsto {f'{}^{\nnl}}\hspace{90pt} {f'{}^{\nnl}}(x)=\fell(\x-\bt).
\end{equation}  
This property is important for several interrelated reasons: 
%\begin{compactenum}
\begin{inparaenum}[(a)]
\item It means that all neurons in a given layer share the same filters (weights). 
\item It implies that no matter where a given object appears in an image, 
the network will treat it the same. 
%treated the same way by the 
%network in the sense that the same set of filters are applied to it in each layer.  
\item Together with the general tendency that higher layers capture larger scale features, it 
helps the CNN learn a meaningful \emph{multiscale} representation of images. 
\item It implies that if we append further layers to the network that are \emph{invariant} to 
translations (such as pooling across all neurons or forming a histogram of their activations), 
then the output of the combined network is guaranteed to be translation invariant. 
\end{inparaenum}
%\end{compactenum}

Recently, \cite{Cohen2016} explored the generalization of equivariance from translations to the action of 
other groups, and \cite{EquivarianceArxiv18} proved that %for even in this generalized setting,    
a generalized form of convolutional (i.e., cross-correlational) structure is a sufficient  
and necessary condition for equivariance to any compact group.  

\subsection{Cross-correlation on the sphere}\label{sec: xcorr}
}
%Section \ref{sec: fourier} explains how such continuous functions can be 
%efficiently represented in Fourier space, and will also shed light on why we allow them 
%to be complex valued. 

\ignore{
Aside from the discretization issue, generalizing cross correlation to the sphere seems relatively straightforward. 
First, we note that the position of any point on the sphere, \m{\x\tin S^2}, can be specified in two different ways: 
in terms of Cartesian coordinates \m{(x_1,x_2,x_3)} (where \m{\nmN{\x}\<=1}) or spherical polars \m{(\theta,\phi)}.  
The inner product between two functions on \m{g,h\colon S^2\to\CC} in terms of these two parametrization is then 
where \m{{}^\ast} denotes the complex conjugate and \m{\mu} is a special measure that ``hides'' 
the cosine factor appearing on the right. 
Further, taking \m{\x_0\<=(0,0)} as the ``origin'' of the sphere (which in Cartesian coordinates is \m{(1,0,0)}),  
we can reach any other point \m{(\theta,\phi)} %on \m{S^2} 
by first applying a rotation \sm{R^{(y)}_\theta} of angle \m{\theta} 
around the \m{y} axis, and then applying a rotation \sm{R^{(z)}_\phi} of angle \m{\phi} around the \m{z} axis.  
Functions can be translated similarly. In particular, given a filter \m{\chi\colon S^2\to\CC}, 
\m{\chi'(\x)=\chi(R^{(y)}^{-1}\x)} is the same filter translated ``Northbound'' by angle \m{\theta}, and 
\m{\chi''(\x)=\chi(R^{(z)}^{-1}\x)} is \m{\chi} translated ``Westbound'' by angle \m{\phi}. 
These formulae then suggest that \rf{eq: xcorr1} can be generalized in the form   
}

\ignore{
One issue with \rf{eq: S^2 convo0} is that it does not take into rotations of \m{\chi} 
around the third coordinate axis, \m{x}. If we are to have a fully \m{\SO(3)}--covariant 
neural architecture, it must treat all three coordinate axes on an equal footing. 
%For small rotation angles, 
This situation is similar to the issue of \emph{steerability} in 2D Vision, 
which has a large literature. We will come back to this issue in Section \ref{sec: gen conv}. 
%i.e. the problem of constructing neural networks or filter banks that are equivariant not just to 
%translations, but also to applying the same filters at different angles. 
%Steerability has a large literature. In the case of discrete rotations, there are relatively simple 
%solutions to the problem. To make a filter bank steerable to rotations by multiples of 90 degrees, 
%it is sufficient to ensure that it has matching horizontal and vertical filters \risi{cite}. 
%Steerability to continuous rotations is a more challenging problem tackled in e.g., \risi{cite}. 
%For an excellent discussion of steerability in the neural networks context, see \cite{Cohen2017}.  

The more serious issue with \rf{eq: S^2 convo0} is that it does not take into account the fact 
that the rotation group \m{\SO(3)} is \emph{not commutative}. 
For example, to compute the match between \m{f} and \m{\chi} when \m{\chi} has been moved to the North Pole \m{(0,0,1)},  
%(i.e., \m{\theta=\pi/2}), 
\rf{eq: S^2 convo0} suggests applying a single rotation \sm{R^{(y)}_{\pi/2}} (Figure \ref{fig: convo1}b). 
However, \m{\chi} could equally well have been moved to \m{(0,0,1)} by 
first rotating it azimuthally by \sm{R^{(z)}_{\pi/2}} and then applying \sm{R^{(y)}_{\pi/2}}. 
Both of these choices map \m{(1,0,0)} to \m{(0,0,1)}, 
but the resulting orientations of \m{\chi} around the pole will differ by 90 degrees, and hence the inner product with \m{f}  
will be different (Figure \ref{fig: convo1}c). 
In this sense, correlation on the sphere, as an operation that takes a pair of functions \m{f,\chi\colon S^2\to\CC}  
and returns a third function \m{\chi\star f\colon S^2\to\CC}, is inherently ill defined. 
}
 
\section{Generalized spherical CNNs}

The starting point for our Generalized Spherical CNNs is the Fourier space correlation 
formula \rf{eq: xcorr Fourier}. 
In contrast to \cite{CohenSpherical18}, however, rather than the geometry, % aspects of this formula, 
we concentrate on its algebraic properties, in particular, its behavior under rotations.
It is well known that if we rotate a spherical function by some \m{R\tin\SO(3)} as 
%a spherical function \m{f} by a rotation \m{R} as given 
in \rf{eq: rotation}, then each vector of its spherical harmonic expansion 
just gets multiplied with the corresponding Wigner D-matrix: 
\begin{equation}\label{eq: covariance1}
\h{f}_\ell\mapsto \rho_\ell(R)\cdot \h{f}_\ell. 
\end{equation}
For functions on \m{\SO(3)}, the situation is similar. If \m{g\colon\SO(3)\to\CC}, and \m{g'} is 
the rotated function \sm{g'(R')=g(R^{-1}R')}, then the Fourier matrices of \m{g'} are 
\m{G'_\ell=\rho_\ell(R)\,G_\ell}. 
The following proposition shows that the matrices output by the \rf{eq: xcorr Fourier} and 
\rf{eq: xcorr Fourier2} cross-correlation formulae behave analogously. 
%, and this is indeed the reason that Spherical CNNs are equivariant. 

\begin{prop}
Let \sm{f\colon S^2\to\CC} be an activation function that under the action of 
a rotation \m{R} transforms as \rf{eq: rotation}, and let \sm{h\colon S^2\to\CC} be a filter. 
Then, each Fourier component of the cross correlation \rf{eq: S^2 xcorr} transforms as 
\begin{equation}\label{eq: covariance2}
\sqbN{\h{h\<\star f}}_\ell\mapsto \rho_\ell(R)\cdot \sqbN{\h{h\<\star f}}_\ell. 
\end{equation}
Similarly, if \sm{f',h'\colon\SO(3)\to\CC}, then \sm{\h{h'\<\star f'}} (as defined in \rf{eq: xcorr Fourier2})
transforms the same way. 
\end{prop}

Equation \rf{eq: covariance1} describes the behavior of spherical harmonic \emph{vectors} under rotations, 
while \rf{eq: covariance2} describes the behavior of Fourier \emph{matrices}. 
However, the latter is equivalent to saying that each column of the matrices separately transforms 
according to \rf{eq: covariance1}.
One of the key ideas of the present paper is to take this property as the basis for the definition of 
covariance to rotations in neural nets. Thus we have the following definition. 
\\

\framedef{
\begin{defn}\label{def: covariant SCNN}
Let \m{\Ncal} be an \m{S\<+1} layer feed-forward neural network whose input is a spherical function 
\m{f^0\colon S^2\to\CC^d}. We say that \m{\Ncal} is a \df{generalized} 
\m{\SO(3)}\df{--covariant spherical CNN} if 
the output of each layer \m{s} can be expressed as a collection of vectors 
\begin{equation}\label{eq: fragments}
\h f^s=(
\underbrace{\h f^{s}_{0,1},\h f^{s}_{0,2},\ldots,\h f^{s}_{0,\tau^s_0}}_{\ell=0},
\underbrace{\h f^{s}_{1,1},\h f^{s}_{1,2},\ldots,\h f^{s}_{1,\tau^s_1}}_{\ell=1},
\ldots\ldots\ldots,
%\h f^{s}_{L,1},\h f^{s}_{L,2},\ldots,
\underbrace{\ldots\h f^{s}_{L,\tau^s_L}}_{\ell=L}
),
\end{equation}
where each \sm{\h f^s_{\ell,j}\tin\CC^{2\ell+1}} is a \m{\rho_\ell}--covariant vector 
in the sense that if the input image is rotated by some rotation \m{R}, then 
\sm{\h f^s_{\ell,j}} transforms as 
\begin{equation}\label{eq: covariance2}
\h f^s_{\ell,j}\mapsto \rho(R)\cdot \h f^s_{\ell,j}.
\end{equation}  
We call the individual \sm{\h f^s_{\ell,j}} vectors the irreducible \df{fragments} of \sm{\h f^s}, 
and the integer vector \m{\tau^s=(\seqz{\tau^s}{L})} counting the number of fragments for each \m{\ell} the 
\df{type} of \sm{\h f^s}. 
\end{defn}
}
\smallskip 

There are a few things worth noting about Definition \ref{def: covariant SCNN}. 
First, since the \rf{eq: covariance2} maps are linear, clearly any \m{\SO(3)}--covariant spherical CNN 
is equivariant to rotations, as defined in the introduction. 
Second, since in \cite{CohenSpherical18} the inputs are functions on the sphere, whereas in higher 
layers the activations are functions on \m{\SO(3)}, their architecture is a special case of 
Definition \ref{def: covariant SCNN} with \m{\tau^0=(1,1,\ldots,1)} and 
\m{\tau^s=(1,3,5,\ldots,2L\<+1)} for \m{s\geq 1}. 

Finally, by the theorem of complete reducibility of representations of compact groups, \emph{any} 
\m{f^s} that transforms under rotations linearily is reducible into a sequence of irreducible fragments 
as in \rf{eq: fragments}. This means that \rf{eq: fragments} is really the most general possible form 
for an \m{\SO(3)} equivariant neural network. 
As we remarked in the introduction, technically, the terms ``equivariant'' and ``covariant'' map to the same 
concept. The difference between them is one of emphasis. We use the term ``equivariant'' when we have 
the same group acting on two objects in a way that is qualitively similar, as in the case 
of the rotation group acting on functions on the sphere and on cross-correlation functions on \m{\SO(3)}. 
We use the term ``covariant'' if the actions are qualitively different, as in the case of rotations of 
functions on the sphere and the corresonding transformations \rf{eq: covariance2} of the irreducible 
fragments in a neural network. 

To fully define our neural network, we need to describe three things: 
1. The form of the linear transformations in each layer involving learnable weights, 
2. The form of the nonlinearity in each layer, 
3. The way that the final output of the network can be reduced to a vector that is rotation 
\emph{invariant}, since that is our ultimate goal. 
The following subsections describe each of these components in turn. 

\subsection{Covariant linear transformations}\label{sec: linear}

In a covariant neural network architecture, the linear operation of each layer must be covariant. 
As described in the Introduction, in classical CNNs, convolution 
automatically satisfies this criterion. In the more general setting of covariance to the 
action of compact groups, the problem was studied in \cite{EquivarianceArxiv18}. 
The specialization of their result to our case is the following. 

\begin{prop}\label{prop: conv1}
Let \sm{\h f^s} be an \m{\SO(3)}--covariant activation function of the form \rf{eq: fragments}, 
and \sm{\h{g}^s=\Lcal(\h f^s)} be a linear function of  \sm{\h f^s} written in a similar form. 
Then \sm{\h{g}^s} is \m{\SO(3)}--covariant if and only each \sm{\h{g}^s_{\ell,j}} fragment 
is a linear combination of fragments from \sm{\h f^s} with the same \m{\ell}. 
\end{prop}

Proposition \ref{prop: conv1} can be made more clear by stacking all fragments of \m{\h f} 
corresponding to \m{\ell} into a \m{(2\ell\<+1)\times \tau^s_\ell} dimensional matrix \m{F^s_\ell}, 
and doing the same for \m{\h g}. Then the proposition tells us simply that 
\begin{equation}\label{eq: conv3}
G^s_\ell=F^s_\ell\,W^s_\ell \hspace{130pt}\ell=0,1,2,\ldots,L
\end{equation} 
for some sequence of complex valued martrices \m{\sseqz{W^s}{L}}. 
Note that \m{W^s_\ell} does not necessarily need to be square, 
i.e., the number of fragments in \sm{\h f} and \sm{\h g} corresponding to \m{\ell} might be different. 
In the context of a neural network, the entries of the \m{W^s_\ell} matrices are learnable parameters. 

Note that the Fourier space cross-correlation formulae \rf{eq: xcorr Fourier} and 
\rf{eq: xcorr Fourier2} are special cases of \rf{eq: conv3} corresponding to taking 
\m{W_\ell\<=\h h_\ell^\dag} or \m{W_\ell\<=H_\ell^\dag}. 
The case of general \m{W_\ell} does not have such an intuitive interpretation in terms of cross-correlation. 
What the \rf{eq: conv3} lacks in interpretability it makes up for in terms of generality, since it 
provides an extremely simple and flexible way of inducing \m{\SO(3)}--covariant linear 
transformations in neural networks.

\subsection{Covariant nonlinearities: the Clebsch--Gordan transform}

Differentiable nonlinearities are essential for the operation of multi-layer neural networks. 
Formulating covariant nonlinearities in Fourier space, however, is more challenging 
than formulating the linear operation. 
This is the reason that most existing group equivariant neural networks perform 
this operation in ``real space''. However, as discussed above, moving back and forth between real space and 
the Fourier domain comes at a signficiant cost and leads to a range of complications involving 
quadrature on the transformation group and numerical errors. 

One of the key contributions of the present paper is to propose a fully Fourier space nonlinearity 
based on the Clebsch--Gordan transform. 
In representation theory, the Clebsch--Gordan decomposition arises in the context of decomposing the 
tensor (i.e., Kronecker) product of irreducible representations into a direct sum of 
irreducibles. In the specific case of \m{\SO(3)}, it takes form 
\begin{equation*}
\rho_{\ell_1}\!(R)\otimes\rho_{\ell_2}\!(R)=C_{\ell_1,\ell_2}\, 
\sqbbigg{~\bigoplus_{\ell=\abs{\ell_1-\ell_2}}^{\ell_1+\ell_2}\rho_\ell(R)~}\: C_{\ell_1,\ell_2}^\top,
\hspace{50pt} R\tin\SO(3),  
\end{equation*}
where \m{C_{\ell_1,\ell_2}} are fixed matrices. 
Equivalently, letting \sm{C_{\ell_1,\ell_2,\ell}} denote the appropriate block of 
columns of \sm{C_{\ell_1,\ell_2}},
\begin{equation*}
\rho_\ell(R)=C_{\ell_1,\ell_2,\ell}^\top\, 
\sqb{\rho_{\ell_1}\!(R)\otimes\rho_{\ell_2}\!(R)}\, C_{\ell_1,\ell_2,\ell}. 
\end{equation*}
The CG-transform is well known in physics, because it is intimately related to the algebra of angular 
momentum in quantum mechanics, %and to multipole expansions, 
and the entries of the \m{C_{\ell_1,\ell_2,\ell}} matrices can be computed relatively easily. 
The following Lemma explains why this construction is relevant to creating 
Fourier space nonlinearities. 

\begin{lemma}\label{lem: CG}
Let \sm{\h f_{\ell_1}} and \sm{\h f_{\ell_2}} be two \sm{\rho_{\ell_1}} resp.~\sm{\rho_{\ell_2}} 
covariant vectors, and \m{\ell} be any integer between \m{\absN{\ell_1\<-\ell_2}} and \m{\ell_1\<+\ell_2}. 
Then 
\begin{equation}\label{eq: CG1}
\h g_\ell=C_{\ell_1,\ell_2,\ell}^\top\sqbbig{\h f_{\ell_1}\otimes \h f_{\ell_2}} 
\end{equation}  
is a \m{\rho_\ell}--covariant vector. 
\end{lemma}

Exploiting Lemma \ref{lem: CG}, the nonlinearity used in our generalized Spherical CNNs consists of 
computing \rf{eq: CG1} between all pairs of fragments. 
In matrix notation, 
\begin{equation}\label{eq: CG2}
G^s_\ell=\bigsqcup_{\abs{\ell_1\<-\ell_2}\leq\ell\leq\ell_1+\ell_2}C_{\ell_1,\ell_2,\ell}^\top 
\sqbbig{F^s_{\ell_1}\otimes F^s_{\ell_2}},   
\end{equation}
where \m{\sqcup} denotes merging matrices horizontally. 
Note that this operation increases the size of the activation substantially: 
the total number of fragments is squared, which can potentially be problematic, and is addressed in the 
following subsection.  

The Clebsch--Gordan decomposition has recently appeared in two preprints discussing   
neural networks for learning physical systems \citep{ThomasTensor18,KondorNbodyArxiv18}. 
However, to the best of our knowledge, in the present context of a general purpose 
nonlinearity, it has never been proposed before. At first sight, the computational cost 
it would appear that the computational cost of \rf{eq: CG1} (assuming that \sm{C_{\ell_1,\ell_2,\ell}} 
has been precomputed) is \m{(2\ell_1\<+1)(2\ell_2\<+1)(2\ell+1)}. 
However, \sm{C_{\ell_1,\ell_2,\ell}} is actually sparse, in particular 
\sm{[C_{\ell_1,\ell_2,\ell}]_{(m_1,m_2),m}=0} unless \m{m_1\<+m_2=m}. Denoting the total number 
of scalar entries in the \sm{{F^s_\ell}_\ell} matrices by \m{N}, this reduces the complexity of 
computing \rf{eq: CG2} to \m{O(N^2 L)}. 
While the CG transform is not currently available as a differentiable operator in any of the major 
deep learning software frameworks, we have developed and will publicly release a C++ PyTorch extension 
for it. 

A more unusual feature of the CG nonlinearity is that its essentially quadratic nature. 
Quadratic nonlinearities are not commonly used in deep neural networks. 
Nonetheless, our experiments indicate that the CG nonlinearity is effective in the context 
of learning spherical images. It is also possible to use higher CG powers, although the 
computational cost will obviously increase. 

\subsection{Limiting the number of channels}

In a covariant network, each individual \m{\h f^s_\ell} fragment is effecively a separate channel. 
In this sense, the quadratic increase in the number of channels after the CG-transform can be 
seen as a natural broadening of the network to capture more complicated features. 
Naturally, allowing the number of channels to increase quadratically in each layer would be 
untenable, though. 

Following the results of Section \ref{sec: linear}, the natural way to counteract the exponential increase 
in the number of channels is follow the CG-transform with another learnable linear transformation that 
reduces the number of fragments for each \m{\ell} to some fixed maximum number \m{\wbar{\tau}_\ell}. 
In fact, this linear transformation can \emph{replace} the transformation of Section \ref{sec: linear}. 
Whereas in conventional neural networks the linear transformation always precedes the nonlinear 
operation, in Clebsch--Gordan networks it is natural to design each layer so as to perform the CG-transform 
first, and then the convolution like step \rf{eq: conv3}, which will limit the number of fragments. 
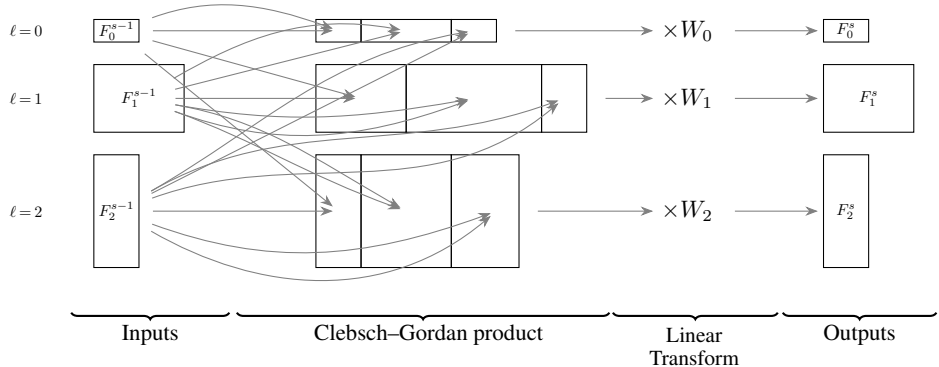
\begin{figure}[t]
\begin{center}
\begin{tikzpicture}[scale=0.3]

\tikzset{arr1/.style={ gray, ->,shorten >=0pt, shorten <=5pt}, >=Stealth}
\tikzset{arr2/.style={ gray, ->,shorten >=0pt, shorten <=5pt}, >=Stealth}

\begin{scope}[xshift=-250, every node/.style={scale=0.6}]
\draw (0,-1) rectangle +(2,-1);
\draw (0,-3) rectangle +(4,-3);
\draw (0,-7) rectangle +(2,-5);
\node(F0) at (1.0,-1.5) {\m{F^{s-1}_0}};
\node(F1) at (2.0,-4.5) {\m{F^{s-1}_1}};
\node(F2) at (1.0,-9.5) {\m{F^{s-1}_2}};
\draw[thick,decoration={brace,mirror},decorate]  (-1.0,-14) to (6,-14);
\node at (2.5,-15) {\Large Inputs}; 

\node at (-3.0,-1.5) {\m{\ell\<=0}}; 
\node at (-3.0,-4.5) {\m{\ell\<=1}}; 
\node at (-3.0,-9.5) {\m{\ell\<=2}}; 
\end{scope}
\begin{scope}[xshift=30, every node/.style={scale=0.6}]

\draw (0,-1) rectangle +(2,-1);
\draw (2,-1) rectangle +(4,-1);
\draw (6,-1) rectangle +(2,-1);
\node(G00) at (1.0,-1.5) {};
\node(G01) at (4.0,-1.5) {};
\node(G02) at (7.0,-1.5) {};

\draw (0,-3) rectangle +(4,-3);
\draw (4,-3) rectangle +(6,-3);
\draw (10,-3) rectangle +(2,-3);
\node(G10) at (2.0,-4.5) {};
\node(G11) at (7.0,-4.5) {};
\node(G12) at (11.0,-4.5) {};

\draw (0,-7) rectangle +(2,-5);
\node(G20) at (1.0,-9.5) {};
\draw (2,-7) rectangle +(4,-5);
\node(G21) at (4.0,-9.5) {};
\draw (6,-7) rectangle +(3,-5);
\node(G22) at (8.0,-9.5) {};

\draw[arr1] (F0) -- (G00);
\draw[arr1] (F0) to [out=20, in=160](G00);
\draw[arr1] (F0) -- (G10);
\draw[arr1] (F1) -- (G10);
\draw[arr1] (F0) -- (G20);
\draw[arr1] (F2) -- (G20);

\draw[arr1] (F1) -- (G01);
\draw[arr1] (F1) to [out=30, in=170](G01);
\draw[arr1] (F2) -- (G02);
\draw[arr1] (F2) to [out=30, in=190](G02);

\draw[arr1] (F1) to [out=-20, in=200] (G11);
\draw[arr1] (F1) to [out=-10, in=190](G11);
\draw[arr1] (F2) to [out=20, in=220] (G12);
\draw[arr1] (F2) to [out=30, in=200](G12);

\draw[arr1] (F1) to [out=-20, in=160] (G21);
\draw[arr1] (F1) to [out=-10, in=150](G21);
\draw[arr1] (F2) to [out=-30, in=220] (G22);
\draw[arr1] (F2) to [out=-20, in=200](G22);

\node(H0) at (8,-1.5) {};
\node(H1) at (12,-4.5) {};
\node(H2) at (9,-9.5) {};
%\node at (0.52,0.5) {\m{u^1}};
\draw[thick,decoration={brace,mirror},decorate]  (-3.5,-14) to (13,-14);
\node at (5.0,-15) {\Large Clebsch--Gordan product}; 
\end{scope}
\begin{scope}[xshift=470, every node/.style={scale=0.9}]
\node(W0) at (1.0,-1.5) {\m{\times W_0}};
\node(W1) at (1.0,-4.5) {\m{\times W_1}};
\node(W2) at (1.0,-9.5) {\m{\times W_2}};

\draw[arr2] (H0) to (W0);
\draw[arr2] (H1) to (W1);
\draw[arr2] (H2) to (W2);

\draw[thick,decoration={brace,mirror},decorate]  (-2.3,-14) to (5,-14);
\node at (1.3,-15) {\small Linear}; 
\node at (1.3,-16) {\small Transform}; 
\end{scope}

\begin{scope}[xshift=670, every node/.style={scale=0.6}]
\draw (0,-1) rectangle +(2,-1);
\draw (0,-3) rectangle +(4,-3);
\draw (0,-7) rectangle +(2,-5);
\node at (1.0,-1.5) {\m{F^s_0}};
\node at (2.0,-4.5) {\m{F^s_1}};
\node at (1.0,-9.5) {\m{F^s_2}};
\node(Fo0) at (0.0,-1.5) {};
\node(Fo1) at (0.0,-4.5) {};
\node(Fo2) at (0.0,-9.5) {};

\draw[arr2] (W0) to (Fo0);
\draw[arr2] (W1) to (Fo1);
\draw[arr2] (W2) to (Fo2);
\draw[thick,decoration={brace,mirror},decorate]  (-1.8,-14) to (5,-14);
\node at (1.6,-15) {\Large Outputs}; 
\end{scope}

\end{tikzpicture}
\end{center}
\caption{\label{fig: layer}
Schematic of a single layer of the Clebsch--Gordan network. 
}
\vspace{-15pt}
\end{figure}

\subsection{Final invariant layer}

After the \m{S\<-1}'th layer, the activations of our network will be a series of matrices 
\sm{\sseqz{F^{S-1}}{L}}, each transforming under rotations according to 
\sm{F^{S-1}_\ell\mapsto \rho_\ell(R)\,F^{S-1}_\ell}. 
Ultimately, however, the objective of the network is to output a vector that is \emph{invariant} 
with respect rotations, i.e., a collection of \emph{scalars}. 
In our Fourier theoretic language, this simply corresponds to the \sm{\h f^S_{0,j}} fragments, since 
the \m{\ell\<=0} representation is constant, and therefore the elements of \sm{F^S_0} are invariant. 
Thus, the final layer can be similar to the earlier ones, except that it only needs to output 
this single (single row) matrix. 

Note that in contrast to other architectures such as \cite{CohenSpherical18} 
that involve repeated forward and backward transforms, 
thanks to their fully Fourier nature, for Clebsch--Gordan nets,  in both training and testing, the elements 
of \sm{F^S_0} are guaranteed to be invariant to rotations of arbitrary magnitude not just 
approximately, but in the \emph{exact} sense, up to limitations of finite precision arithmetic. 
This is a major advantage of Clebsch--Gordan networks compared to other covariant architectures. 

\subsection{Summary of algorithm} \label{sec: summary}

In summary, our Spherical Clebsch--Gordan network is an \m{S\<+1} layer feed-forward neural network 
in which  apart from the initial spherical harmonic transform, every other operation is a simple 
matrix operation. In the forward pass, the network operates as follows. 

\begin{enumerate}[~1.]
	\item The inputs to the network are \m{n_{\textrm{in}}} functions 
	\m{\sseq{f^0}{n_\textrm{in}}\!\colon\! S^2\<\to\CC}. 
	For example, for spherical color images, \sm{f^0_1,f^0_2} and \m{f^0_3} might encode the red, green and blue 
	channels. For generality, we treat these functions as complex valued, but of course they may also be real.
	%\item 
	The activation of layer \m{s\<=0} is the union of the spherical transforms of these functions 
	\m{\sseq{f^0}{n_\textrm{in}}} 
	up to some band limit \m{L}, i.e., %\sm{\h f^0_{\ell,j}=[\wt f^0_j]_\ell^m} with 
	%\sm{\h f^0_{\ell,j}=([f^0_{\ell,j}]_{-\ell},\ldots,[f^0_{\ell,j}]_\ell)^\top} with 
	\begin{equation}
	[\h f^0_{\ell,j}]_{m}=
	\ovr{4\pi}\int_{0}^{2\pi} \int_{-\pi}^{\pi}\! f^0_{j}(\theta,\phi)^\ast\:Y_\ell^m(\theta,\phi)\tts\cos\theta\ts d\theta d\phi. 
	\end{equation}
	Therefore, the type of \m{\h f^0} is \m{\tau^0\<=(n_{\textrm{in}},n_{\textrm{in}},\ldots,n_{\textrm{in}})}, and 
	\m{\h f^0} is stored as a collection of \m{L\<+1} matrices \m{\cbrN{F^0_0,F^0_1,\ldots,F^0_L}} of sizes 
	\m{1\<\times n_{\textrm{in}}}, \m{3\<\times n_{\textrm{in}}}, \m{5\<\times n_{\textrm{in}}},\ldots % etc..  
	\item For layers \m{s=1,2,\ldots,S\<-1}, the Fourier space activation \m{\h f^s\<=\brN{F^s_0,F^s_1,\ldots,F^s_L}} 
	is computed as follows: 
	\begin{enumerate}
		\item %Tensor products are \m{\cbrN{F^{s-1}_0,F^{s-1}_1,\ldots,F^{s-1}_L}}
		We form all possible Kronecker products 
		\begin{equation}
		G^s_{\ell_1,\ell_2}=F^{s-1}_{\ell_1}\otimes F^{s-1}_{\ell_2}
		%\tin \CC^{(2\ell_1+1)(2\ell_2+1)\times \tau^{s-1}_{\ell_1}\tau^{s-1}_{\ell_2}}
		\qqquad 0\leq \ell_1\leq \ell_2\leq L.
		\end{equation}
		Note that the size of \m{G^s_{\ell_1,\ell_2}} is \sm{{(2\ell_1+1)(2\ell_2+1)\times (\tau^{s-1}_{\ell_1}\tau^{s-1}_{\ell_2})}}.
		\item Each \m{G^s_{\ell_1,\ell_2}} is decomposed into \m{\rho_\ell}--covariant blocks by 
		\begin{equation}
		[G^s_{\ell_1,\ell_2}]_\ell=C^\dag_{\ell_1,\ell_2,\ell}\: G^s_{\ell_1,\ell_2},
		\end{equation}
		where \m{C^\dag_{\ell_1,\ell_2,\ell}} is the inverse Clebsch--Gordan matrix as in \rf{eq: CG1}. 
		\item All \sm{[G^s_{\ell_1,\ell_2}]_\ell} blocks with the same \m{\ell} are concatenated into a 
		large matrix \m{H^s_\ell\tin\CC^{(2\ell\<+1)\times \wbar{\tau^s_\ell}}}, 
		and this is multiplied by the weight matrix \m{W^s_\ell\tin \CC^{\wbar{\tau^s_\ell}\times\tau^s_\ell}} 
		to give
		\begin{equation}
		F^s_\ell=H^s_\ell\,W^s_\ell\qqquad \qqquad \ell=0,1,\ldots,L.
		\end{equation}
	\end{enumerate}
	\item The operation of layer \m{S} is similar, except that the output type is 
	\m{\tau^S=(n_{\textrm{out},0,0,\ldots,0})}, so components with \m{\ell>0} do not need to be 
	computed. 
	By construction, the %\m{n_{\textrm{out}}} 
	entries of \sm{F^s_0\tin\CC^{1\times n_{\textrm{out}}}} are \m{\SO(3)}--invariant scalars, i.e., 
they are invariant to the simulatenous rotation of the \m{\sseq{f^0}{n_\textrm{in}}} inputs. 
These scalars may be passed on to a fully connected network or plugged directly into a loss function.  
\end{enumerate}

\noindent 
The learnable parameters of the network are the \sm{\cbrN{W^s_\ell}_{s,\ell}} weight matrices. 
The matrices are initialized with random complex entries. 
As usual, the network is trained by backpropagating the gradient of the loss from layer \m{S} 
to each of the weight matrices. 

\section{Experiments}

In this section we describe experiments that give a direct comparison with those reported by Cohen \emph{et al.} \citep{CohenSpherical18}. We choose these experiments as the Spherical CNN proposed in \citep{CohenSpherical18} is the only direct competition to our method. Besides, the comparison is also instructive for two different reasons: Firstly, while the procedure used in \citep{CohenSpherical18} is exactly equivariant in the discrete case, for the continuous case they use a discretization which causes their network to partially lose equivariance with changing bandwidth and depth, whereas our method is always equivariant in the exact sense. Secondly, owing to the nature of their architecture and discretization, \citep{CohenSpherical18} use a more traditional non-linearity i.e. the ReLU, which is also quite powerful. In our case, to maintain full covariance and to avoid the quadrature, we use an unconventional quadratic non-linearity in fourier space. Because of these two differences, the experiments will hopefully demonstrate the advantages of avoiding the quadrature and maintaining full equivariance despite using a purportedly weaker nonlinearity.

\paragraph{Rotated MNIST on the Sphere} \label{sec - MNIST}
We use a version of MNIST in which the images are painted onto a sphere and use two instances as in \citep{CohenSpherical18}: One in which the digits are projected onto the northern hemisphere and another in which the digits are projected on the sphere and are also randomly rotated.

The baseline model is a classical CNN with 5 $\times$ 5 filters and 32, 64, 10 channels with a stride of 3 in each layer (roughly 68K parameters). This CNN is trained by mapping the digits from the sphere back onto the plane, resulting in nonlinear distortions. The second model we use to compare to is the Spherical CNN proposed in \citep{CohenSpherical18}. For this method, we use the same architecture as reported by the authors i.e. having layers $S^2$ convolution -- ReLU -- $SO(3)$ convolution -- ReLU -- Fully connected layer with bandwidths 30, 10 and 6, and the number of channels being 20, 40 and 10 (resulting in a total of 58K parameters). 

For our method we use the following architecture:  We set the bandlimit $L_{max} = 10$,  and keep $\tau_l = \ceil{\frac{12}{\sqrt{2 l + 1}}}$, using a total of 5 layers as described in section 3.5, followed by a fully connected
layer of size 256 by 10. We use a variant of batch normalization that preserves covariance in the Fourier layers. This method takes a expanding average of the standard deviation for a particular fragment for all examples seen during training till
then and divide the fragment by it (in testing, use the average from training); the parameter corresponding to the mean in usual batch normalization is kept to be zero as anything else will break covariance. Finally, we concatenate the output of each $F_0^s$ in each internal layer (length 24 each, as each is $\tau_0 = 12$ complex numbers), as well as the original coefficient at $l=0$ (length 2), into a $SO(3)$ invariant vector of length 122. (We observed that having these skip connections was crucial to facilitate smooth training.) After that, we use the usual batch normalization \cite{IoffeSzegedy} on the concatenated results before feeding it into fully connected layers of length 256, a dropout layer with dropout probability 0.5, and finally a linear layer to to 10 output nodes. The total number of parameters was 285772, the network was trained by using the ADAM optimization procedure \cite{adam2015} with a batch size of 100 and a learning rate of $5\times10^{-4}$. We also
used L2 weight decay of $1 \times10^{-5}$ on the trainable parameters.

We report three sets of experiments: For the first set both the training and test sets were not rotated (denoted NR/NR), for the second, the training set was not rotated while the test was randomly rotated (NR/R) and finally when both the training and test sets were rotated (denoted R/R).

%\ignore{
\begin{center}
{\small 
\begin{tabular}{l*{3}{c}}
	Method              & NR/NR & NR/R & R/R \\
	\hline \hline
	Baseline CNN & 97.67 & 22.18 & 12  \\
	Cohen \emph{et al.} & 95.59 & 94.62 & 93.4  \\
	Ours (FFS2CNN)  & 96.4 & 96 & 96.6  \\
\end{tabular}
}
\end{center}
%}

We observe that the baseline model's performance deteriorates in the three cases, more or less reducing to random chance in the R/R case. While our results are better than those reported in \citep{CohenSpherical18}, they also have another characteristic: they remain roughly the same in the three regimes, while those of \citep{CohenSpherical18} slightly worsen. We think this might be a result of the loss of equivariance in their method. 

\paragraph{Atomization Energy Prediction}
Next, we apply our framework to the QM7 dataset \citep{QM71, QM72}, where the goal is to regress over atomization energies of molecules given atomic positions ($p_i$) and charges ($z_i$). Each molecule contains up to 23 atoms of 5 types (C, N, O, S, H). More details about the representations used, baseline models, as well as the architectural parameters are provided in the appendix. 
We use the Coulomb Matrix (CM) representation proposed by \citep{QM72}, which is rotation and translation invariant but not permutation invariant. The Coulomb matrix $C \in \mathbb{R}^{N \times N}$ is defined such that for a pair of atoms $i \neq j$, $C_{ij} = (z_iz_j)/(|p_i - p_j|)$, which represents the Coulomb repulsion, and for atoms $i=j$, $C_{ii} = 0.5 z_i^{2.4}$, which denotes the atomic energy due to charge. To test our algorithm we use the same set up as in \citep{CohenSpherical18}: We define a sphere $S_i$ around $p_i$ for each atom $i$. Ensuring uniform radius across atoms and molecules and ensuring no intersections amongst spheres during training, we define potential functions $U_z(x) = \sum_{j\neq i, z_j = z} \frac{zi z}{|x - p_i|}$ for every $z$ and for every $x$ on $S_i$. This yields a $T$ channel spherical signal for each atom in a molecule. This signal is then discretized using Driscol-Healy \citep{DriscollHealy} grid using a bandwidth of $b = 10$. This gives a sparse tensor representation of dimension $N \times T \times 2b \times 2b$ for every molecule.

Our spherical CNN architecture has the same parameters and hyperparameters as in the previous subsection except that $\tau_l = 15$ for all layers, increasing the number of parameters to 1.1 M. Following \citep{CohenSpherical18}, we share weights amongst atoms and each molecule is represented as a $N \times F$ tensor where $F$ represents \sm{F^s_0} scalars concatenated together. Finally, we use the approach proposed in \citep{Zaheer2017} to ensure permutation invariance. The feature vector for each atom is projected onto 150 dimensions using a MLP. These embeddings are summed over atoms, and then the regression target is trained using another MLP having 50 hidden units. Both of these MLPs are jointly trained. The final results are presented below, which show that our method outperforms the Spherical CNN of Cohen \emph{et al.}. The only method that delivers better performance is a MLP trained on randomly permuted Coulomb matrices \citep{Montavon2012}, and as \citep{CohenSpherical18} point out, this method is unlikely to scale to large molecules as it needs a large sample of random permutations, which grows rapidly with $N$.

\begin{center}
	\begin{tabular}{l*{2}{c}}
		Method  &  RMSE \\
		\hline \hline
		MLP/Random CM \citep{Montavon2012} & \bf{5.96}\\
		\hline
		LGIKA (RF) \citep{Raj2016} & 10.82\\
		\hline
		RBF Kernels/Random CM \citep{Montavon2012} & 11.42\\
		\hline
		RBF Kernels/Sorted CM \citep{Montavon2012} & 12.59\\
		\hline
		MLP/Sorted CM \citep{Montavon2012} & 16.06\\
		\hline
		Spherical CNN \citep{CohenSpherical18} &  8.47  \\
		\hline
		Ours (FFS2CNN)& \bf{7.97} \\
	\end{tabular}
\end{center}

\paragraph{3D Shape Recognition}
Finally, we report results for shape classification using the SHREC17 dataset \citep{SHREC}, which is a subset of the larger ShapeNet dataset \citep{ShapeNet} having roughly 51300 3D models spread over 55 categories.  It is divided into a 70/10/20 split for train/validation/test. Two versions of this dataset are available: A regular version in which the objects are consistently aligned and another where the 3D models are perturbed by random rotations. Following \citep{CohenSpherical18} we focus on the latter version, as well as represent each 3D mesh as a spherical signal by using a ray casting scheme. For each point on the sphere, a ray towards the origin is sent which collects the ray length, cosine and sine of the surface angle. In addition to this, ray casting for the convex hull of the mesh gives additional information, resulting in 6 channels. The spherical signal is discretized using the Discroll-Healy grid \citep{DriscollHealy} with a bandwidth of 128. We use the code provided by \citep{CohenSpherical18} for generating this representation. 

We use a ResNet style architecture, but with the difference that the full input is not fed back but rather different frequency parts of it. We consider $L_{max} = 14$, and first train a block only till $L = 8$ using $\tau_l = 10$ using 3 layers. The next block consists of concatenating the fragments obtained from the previous block and training for two layers till $L = 10$, repeating this process till $L_{max}$ is reached. These later blocks use $\tau_l = 8$.  As earlier, we concatenate the \sm{F^s_0} scalars from each block to form the final output layer, which is connected to 55 nodes forming a fully connected layer. We use Batch Normalization in the final layer, and the normalization discussed in \ref{sec - MNIST} in the Fourier layers. The model was trained with ADAM using a batch size of 100 and a learning rate of $5 \times 10^{-4}$, using L2 weight decay of 0.0005 for regularization. We compare our results to some of the top performing models on SHREC (which use architectures specialized to the task) as well as the model of Cohen \emph{et al.}. Our method, like the model of Cohen \emph{et al.} is task agnostic and uses the same representation. Despite this, it is able to consistently come second or third in the competition, showing that it affords an efficient method to learn from spherical signals. 

\begin{center}
	\begin{tabular}{l*{5}{c}}
		Method              & P@N & R@N & F1@N & mAP & NDCG \\
		\hline \hline
		Tatsuma\_ReVGG & 0.705 & 0.769 & 0.719 & 0.696 & 0.783  \\
		Furuya\_DLAN & 0.814 & 0.683 & 0.706 & 0.656 & 0.754  \\
		SHREC16-Bai\_GIFT  & 0.678 & 0.667 & 0.661 & 0.607 & 0.735  \\
		Deng\_CM-VGG5-6DB  & 0.412 & 0.706 & 0.472 & 0.524 & 0.624  \\
		Spherical CNNs \citep{CohenSpherical18}  & 0.701 & 0.711 & 0.699 & 0.676 & 0.756  \\
		\hline
		FFS2CNNs (ours)  & 0.707 & 0.722 & 0.701 & 0.683 & 0.756  \\
	\end{tabular}
\end{center}
 
\section{Conclusion}

We have presented an $\SO(3)$-equivariant neural network architecture for spherical data that operates 
completely in Fourier space, circumventing a major drawback of earlier models 
that need to switch back and forth between Fourier space and ``real'' space. 
% for the convolution operation and applying the non-linearity respectively. 
We achieve this by -- rather unconventionally -- using the Clebsch-Gordan decomposition 
as the only source of nonlinearity. 
While the specific focus is on spheres and $\SO(3)$-equivariance, the approach is more widely applicable, 
suggesting a general formalism for designing fully Fourier neural networks that are equivariant to 
the action of any compact continuous group. 

\clearpage

{\small
\setlength{\bibsep}{2pt}
\bibliographystyle{unsrt}
\bibliography{SphericalCNN-NIPS18}
}

\end{document}